\documentclass{gretsi}

\usepackage[latin1]{inputenc}
\usepackage[french,english]{babel}
\usepackage{graphicx,afterpage}
\usepackage{fancybox}
\usepackage{psboxit,pstricks,rotating}
\usepackage{psfrag}
\usepackage{ulem}
\usepackage{amsmath,bbm} 
\usepackage{amsfonts} 
\usepackage{amssymb}
\usepackage{vector}
\usepackage{floatflt}
\usepackage{theorem}
\usepackage{stmaryrd}
\usepackage{euscript}
\usepackage{pifont}
\usepackage{subfigure}
\usepackage{array}
\usepackage[T1]{fontenc}
\usepackage{endnotes}

\title{IR image databases generation under target intrinsic thermal variability constraints}
\auteur{\coord{J\'er\^ome}{Gilles}{1}, \coord{St\'ephane}{Landeau}{2}, \coord{Tristan}{Dagobert}{2}, \coord{Philippe}{Chevalier}{3}, \coord{Christian}{Bolut}{3}}
\adresse{\affil{1}{DGA-CEP/EORD, 16bis rue Prieur de la C\^ote d'Or 94110 Arcueil}\affil{2}{DGA-CEP/LOT, 16bis rue Prieur de la C\^ote d'Or 94110 Arcueil}\affil{3}{DGA-ETAS, BP 60036, Montreuil-Juign\'e 49245 Avrill� Cedex}}
\email{jerome.gilles@etca.fr, stephane.landeau@dga.defense.gouv.fr,\\tristan.dagobert@dga.defense.gouv.fr, philippe.chevalier@dga.defense.gouv.fr}

\englishabstract{This paper deals with the problem of infrared image database generation for ATR assessment purposes. Huge databases are required to have quantitative and objective performance evaluations. We propose a method which superimpose targets and occultants on background under image quality metrics constraints to generate realistic images. We also propose a method to generate target signatures with intrinsic thermal variability based on 3D models plated with real infrared textures.}

\begin{document}

\maketitle

\section{Introduction}
The evaluation and the parameter optimization of an ATR algorithm in thermal IR imagery are basically dependent on the quality and the availability of usable image databases.

The acquisition of this kind of data\-ba\-ses presents a relatively high cost and takes a lots of time. A solution holds in the use of scenes simulators, but those remain expensive in times computing and it is especially difficult to select the various parameters in order to sweep a maximum of operational scenarios in an exhaustive way.

We propose to generate hybrid databases by superimposition of targets and occultant in front of a background under constraint of image quality metrics. Entry parameters of these constraints are most effective to describe realistic operational scenarios. Moreover, one important aspect of the IR imagery is the intrinsic thermal variability of the target signature. We propose in this paper an original method allowing to take into account of this variability during the scene generation. For that, we use real images of each vehicle acquired in their ``extremes'' operating process: target at ambient temperature, target with all its potentially hot elements to the maximum of their temperatures. These signatures are then plated on a 3D model of the vehicle segmented under subelements of signature considered to have an homogeneous and independent thermal behavior. It is thus possible to selectively parameterize the temperature of these under-parts, to build alternatives of the signature of the same target. We use a 2D projection of this model under the wished visual angle for finally superimposing it under constraint and applying the wanted sensor effect to it.

In section \ref{sec:hyb}, we start by recalling the hybrid scene generation principle proposed in previous work \cite{landeau}. In section \ref{sec:var}, we give a complete overview of the method which permits us to create vehicle signature with intrinsic thermal variability. In section \ref{sec:res}, we give some results obtained by the proposed method and we conclude in section \ref{sec:con}.

\section{Hybrid scenes generation}\label{sec:hyb}
\begin{figure}[t]
\begin{center}
\includegraphics[width=0.45\textwidth]{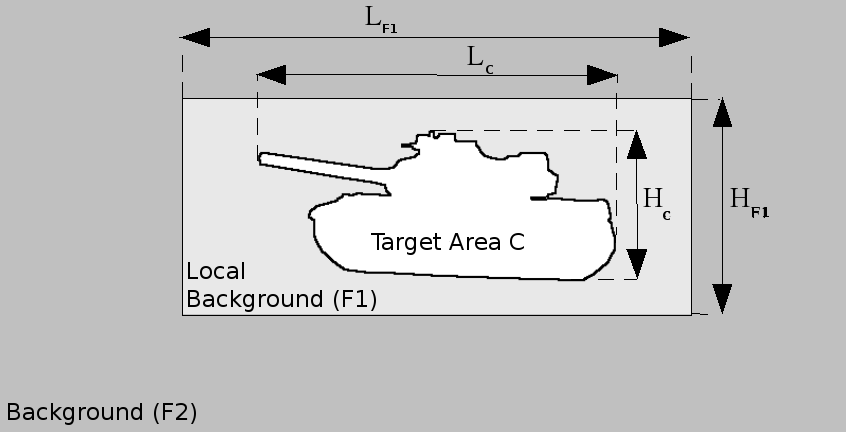}
\end{center}
\caption{Definition of different areas for target superimposition over a choosen background.}
\label{fig:zones}
\end{figure}

\begin{figure}[t]
\begin{center}
\includegraphics[width=0.45\textwidth]{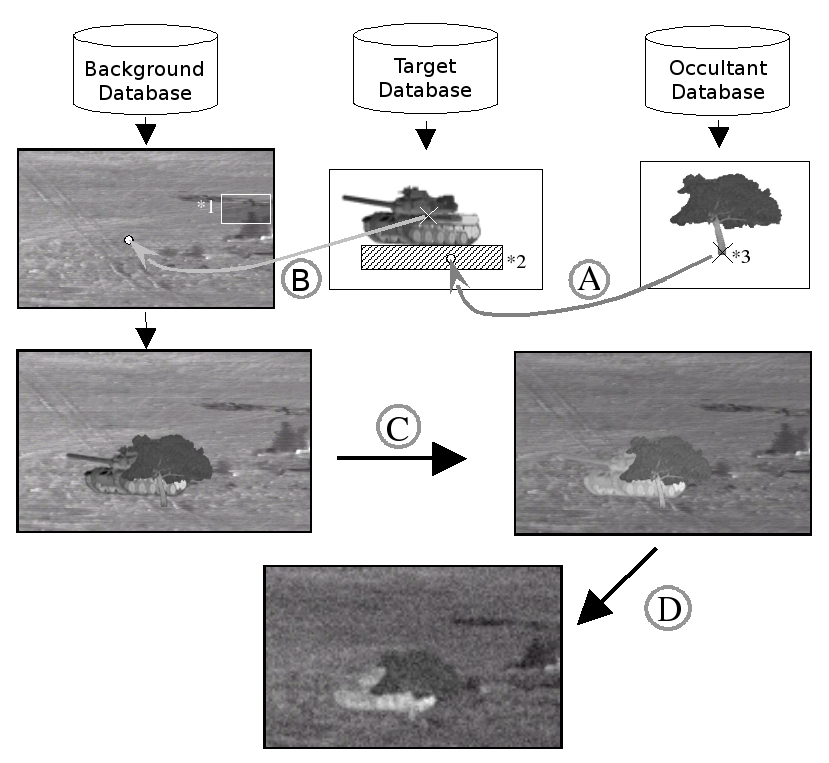}
\end{center}
\caption{Hybrid scene generation principle.}
\label{fig:gene}
\end{figure}

In this section, we recall the hybrid scene generation principle proposed in \cite{landeau}. This generation is called ``hybrid'' because it consists to superimpose a real target signature, eventually with different kind of occultants (like trees, rocks,$\ldots$), in a real background. The interest of the method is that it is possible to control the output image quality by some metrics \cite{driggers,nvesd}. The used metrics are: local contrast $RSS$, ``detectability'' quantity $Q_D$, signal to clutter ratio $SCR$, occultation ratio $R_x$ and internal target contrast $K$. These quantities are defined by

\begin{align}
RSS&=\frac{1}{\nu_k}\sqrt{(\mu_{C}-\mu_{F_1})^2+\sigma_{C}^2} \\
Q_D&=RSS.S_{C} \\
SCR&=\frac{\nu_kRSS}{\sigma_F} \\
R_x&=\frac{S_{\text{occluded target area}}}{S_{\text{full target area}}} \\
K&=\frac{\mu_{F_1}-\mu_{C}}{\nu_kRSS}=\frac{\Delta\mu}{\nu_kRSS}
\end{align}

where $C$ is the target, $F_1$ the local background over $C$ and $F_2$ the remaining background (we denote the global background $F=F_1\cup F_2$), see Figure \ref{fig:zones}. The quantities $S_x$, $\mu_x$, $\sigma_x$ are the surface, mean and standard deviation of $x$ area where $x$ is $C$, $F_1$ or $F_2$, respectively. The coefficient $\nu_k$ is the coefficient which permits to do the conversion between pixel gray levels and temperature in Kelvin. The choice of these parameters fixes some gains and offsets to apply on the pixels of both the target and background in order to obtain the resulting image. Finally, the sensor effect (MTF and noise) is applied. The hybrid scene generation process is summarised in Figure \ref{fig:gene}. We start, \textcircled{\footnotesize{A}}, by positioning the occultant, then, \textcircled{\footnotesize{B}}, the positioning of the target inside the background. We apply the calculated gains and offsets to histograms of each region, \textcircled{\footnotesize{C}}. We finish by applying the sensor effect, \textcircled{\footnotesize{D}}. More details and the expressions of the different gains and offsets to apply can be found in \cite{landeau}. This scene generation principle was used for ATR algorithms evaluation in the CALADIOM project. However, one aspect is not taken into account in this algorithm: the intrinsic thermal variability of targets. In the next section, we propose an approach to deal with this aspect in the hybrid scene generation.

\section{Intrinsic thermal variability of targets}\label{sec:var}
In this section, we propose a new method which permits to deal with the intrinsic thermal variability of a target in IR imagery. Indeed, a same target can have many different thermal aspects according to its activity. For example, tires of a vehicle which stopped since a long time are more cold than a vehicle which is moving for a long time. But their engine are quite identical. However, the current vehicle recognition algorithms use some training to be efficient. It is obvious that this kind of algorithm will have less performances if they never ``learn'' the different aspects of concerned targets.\\

\begin{figure}[t]
\begin{center}
\includegraphics[width=0.35\textwidth]{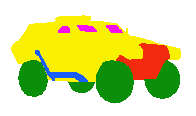}
\end{center}
\caption{Map of homogeneous thermal behavior of a given vehicle.}
\label{fig:decoup}
\end{figure}

This variability is function of the vehicle operation, this means that it is sufficient to modify the signature of the vehicle. As it is too complex, from a practical point of view, to use an accurate thermal physical models for different targets, we propose to create intermediate signatures by interpolation from ambient ($TA$) and operationnal ($TF$) temperatures, taken from real radiometric images from ETAS. For that, we lay out 3D models of vehicles on which we plate infrared textures. These textures are available for the $TA$ and $TF$ temperatures. We propose to segment the surface of the vehicle into homogeneous thermal behavior areas which are dependent on the different operationnal vehicle's areas. The relevant choosen areas are: the engine, the main body, the muffler, windows, tires/caterpillar (see Figure \ref{fig:decoup}).

An intermediate thermal state of an area $R$, relevant to the wanted variability, is generated by mixing the states $TA$ and $TF$, according to equation \ref{eq:mix}.

\begin{equation}\label{eq:mix}
TI_R=(1-\lambda_R)TA_R+\lambda TF_R,
\end{equation}
where $\lambda\in [0;1]$ represent the variability rate. We can define three different behavior:
\begin{enumerate}
\item ambient temperature: $\lambda\in[0;0.1]$,
\item intermediate temperature: $\lambda\in]0.1;0.9[$,
\item in operation temperature: $\lambda\in[0.9;1]$.
\end{enumerate}

\begin{figure}[t]
\begin{center}
\includegraphics[width=0.45\textwidth]{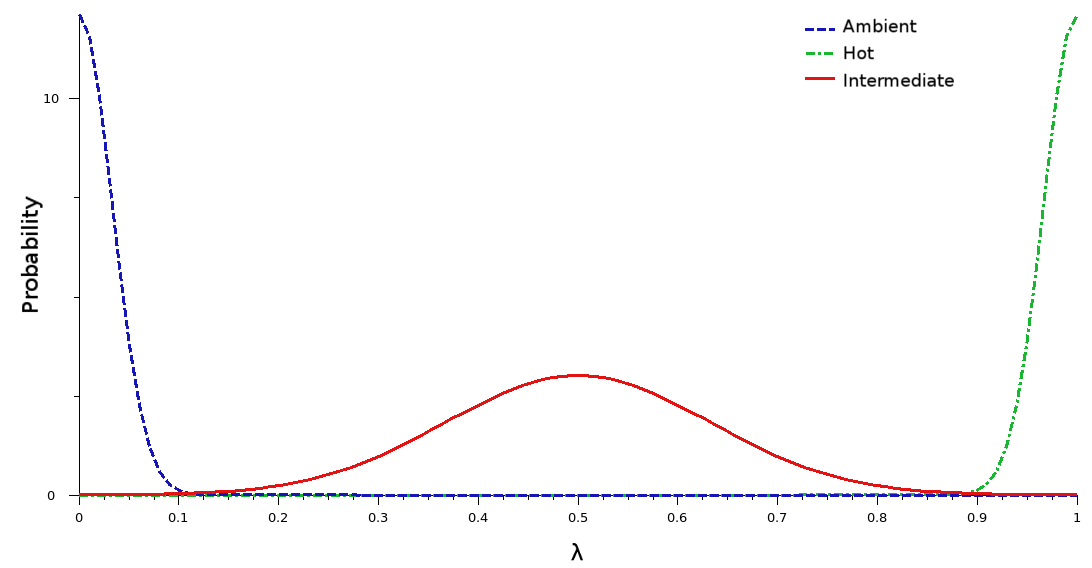}
\end{center}
\caption{Probability laws of $\lambda$ for the different operationnal mode.}
\label{fig:gauss}
\end{figure}

The final choice of $\lambda$ is done by random drawings according to gaussian laws (or half-gaussian at extremities, see Figure \ref{fig:gauss}). The standard deviation of each gaussian is choosen in order to have $99\%$ of its surface inside the intervals considered above. This is equivalent to $3\sigma_{TA}=3\sigma_{TF}=0.1$ and $3\sigma_{TI}=0.4$, this give us $\sigma_{TA}=\sigma_{TF}=0.33$ and $\sigma_{TI}=0.133$ respectively. Then the different laws are given by equations (\ref{eq:pta}), (\ref{eq:ptf}) and (\ref{eq:pti}) (for all $\lambda$ taken in the previous intervals).

\begin{align}
P_{TA}(\lambda)&=\frac{1}{\sqrt{2\pi\sigma_{TA}^2}}\exp(-\frac{\lambda^2}{2\sigma_{TA}^2}) \label{eq:pta} \\
P_{TF}(\lambda)&=\frac{1}{\sqrt{2\pi\sigma_{TF}^2}}\exp(-\frac{(1-\lambda)^2}{2\sigma_{TF}^2}) \label{eq:ptf} \\
P_{TI}(\lambda)&=\frac{1}{\sqrt{2\pi\sigma_{TI}^2}}\exp(-\frac{(\lambda-0.5)^2}{2\sigma_{TI}^2}) \label{eq:pti}
\end{align}

By selecting different thermal configuration (for example a vehicle in stand by where its engine and muffler are hot, its body, windows and tires at ambient temperature), we can generate the intermediate texture to plate on the 3D model. Thus, we can make 2D views, at different angles of view, by projection in order to increase the number of signatures in the database used for hybrid scene generation.

\section{Results}\label{sec:res}
In this section, we show some results we get by the previously described method. 

\begin{figure}[t]
\begin{tabular}{cc}
\includegraphics[width=0.22\textwidth]{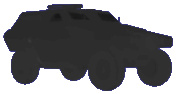} & \includegraphics[width=0.22\textwidth]{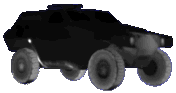} \\
\includegraphics[width=0.22\textwidth]{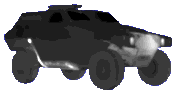} & \includegraphics[width=0.22\textwidth]{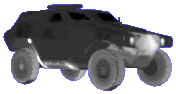}
\end{tabular}
\caption{Example of different thermal configurations generated by the proposed methode.}
\label{fig:sign}
\end{figure}

First, Figure \ref{fig:sign} shows different thermal configurations of a same vehicle presented in the same point of view. We can see that it is possible to create realistic IR signatures which correspond to predefined operationnal states (vehicle completely motionless, vehicle in motion, $\ldots$). In conclusion, the method enables us the generation of all needed views of a vehicle.

\begin{figure}[t]
\begin{center}
\includegraphics[width=0.4\textwidth]{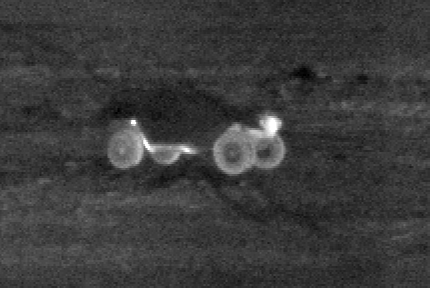} 
\includegraphics[width=0.4\textwidth]{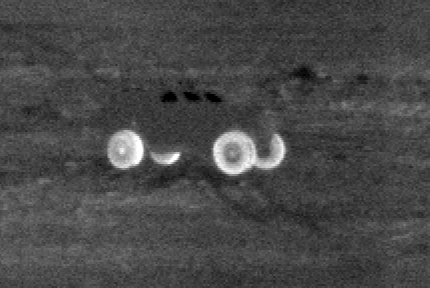} 
\includegraphics[width=0.4\textwidth]{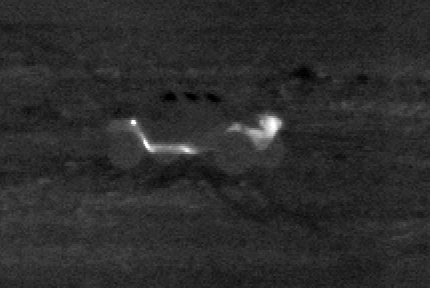} 
\caption{Example of a same scene with different thermal configurations of a vehicle.}
\label{fig:incr}
\end{center}
\end{figure}

Second, these new signatures are added to a new target database which will be used by the hybrid scene generator. This allows us to generate scenes which hold account the intrinsic thermal variability of targets by superimposing the wanted target taken in this new database. Figure \ref{fig:incr} shows an example of a same scene, generated with the same image quality constraints, containing the same vehicle with different thermal configurations.

\section{Conclusion}\label{sec:con}
In this paper, we present a new methodology which permits to generate realistic infrared scene images for ATR learning and assessment purposes. The method superimpose IR signature of vehicle in a scene background, eventually by adding some occultants, under image quality constraints. We also propose to extend the method to deal with the possibility of adding intrinsic thermal variability of targets. To do this, we interpolate an intermediate signature from ambient and hot signatures. This simulated signature is then plated on a 3D model which is projected from a desired point of view.

This method permits us to build image databases to tune and to evaluate ATR algorithms. The advantage of this method is the easyness of building huge databases and furthermore to have quantitative measures of algorithm performances. Today, a database of more ten thousand images, with their groundtruth, was distributed to the research teams.

We currently extend the method in order to generate images sequences with moving targets. The principle is to provide an a priori trajectory. Then the algorithm calculate the necessary different angle of view based on the sensor position in the scene. The knowledge of these angles permits it to select in the database the right signature to superimpose.

\end{document}